\documentclass[10pt,twocolumn,letterpaper]{article}

\usepackage{iccv}
\usepackage{times}
\usepackage{epsfig}
\usepackage{graphicx}
\usepackage{amsmath}
\usepackage{amssymb}
\usepackage{caption}
\usepackage{subcaption}
\DeclareMathOperator*{\argmax}{argmax}

\usepackage[breaklinks=true,bookmarks=false]{hyperref}

\iccvfinalcopy 

\def\iccvPaperID{****} 

\ificcvfinal\fi

\begin{document}

\title{CLTA: Contents and Length-based Temporal Attention for Few-shot Action Recognition}

\author{Yang Bo \hspace{2cm} Yangdi Lu \hspace{2cm} Wenbo He \\
McMaster University\\
{\tt\small boy2@mcmaster.ca \hspace{.3cm} \tt\small luy100@mcmaster.ca \hspace{.3cm} \tt\small hew11@mcmaster.ca}
}

\maketitle
\ificcvfinal\thispagestyle{empty}\fi

\begin{abstract}
Few-shot action recognition has attracted increasing attention due to the difficulty in acquiring the properly labelled training samples. Current works have shown that preserving spatial information and comparing video descriptors are crucial for few-shot action recognition. However, the importance of preserving temporal information is not well discussed. In this paper, we propose a Contents and Length-based Temporal Attention (CLTA) model, which learns customized temporal attention for the individual video to tackle the few-shot action recognition problem. CLTA utilizes the Gaussian likelihood function as the template to generate temporal attention and trains the learning matrices to study the mean and standard deviation based on both frame contents and length. We show that even a not fine-tuned backbone with an ordinary softmax classifier can still achieve similar or better results compared to the state-of-the-art few-shot action recognition with precisely captured temporal attention.
\end{abstract}

\section{Introduction} \label{introuction}
The performance of deep learning models on visual recognition tasks heavily relies on abundant labelled training instances. However, it is error-prone and labour-intensive to obtain the labelled training samples. Consequently, the problem of classifying unseen classes with few examples, known as few-shot classification, has attracted considerable attention. The majority of recent few-shot learning efforts focus on image classification. Examples include model initialization-based methods~\cite{ravi2016optimization, finn2017model}, metric learning-based methods~\cite{vinyals2016matching, snell2017prototypical, sung2018learning}, gradient-based methods~\cite{rusu2018meta, simon2020modulating, zintgraf2019fast} and hallucination-based methods~\cite{hariharan2017low, zhang2019few}. For video data, the few-short classification is more required but more challenging. It is even harder to obtain correct labelled videos than images, especially for the tasks which need precise boundaries (e.g. multi-activities detection). It is hard to say the action starts
and ends at a specific frame. 
\begin{figure*}[!th]
\begin{center}
\includegraphics[width=\textwidth]{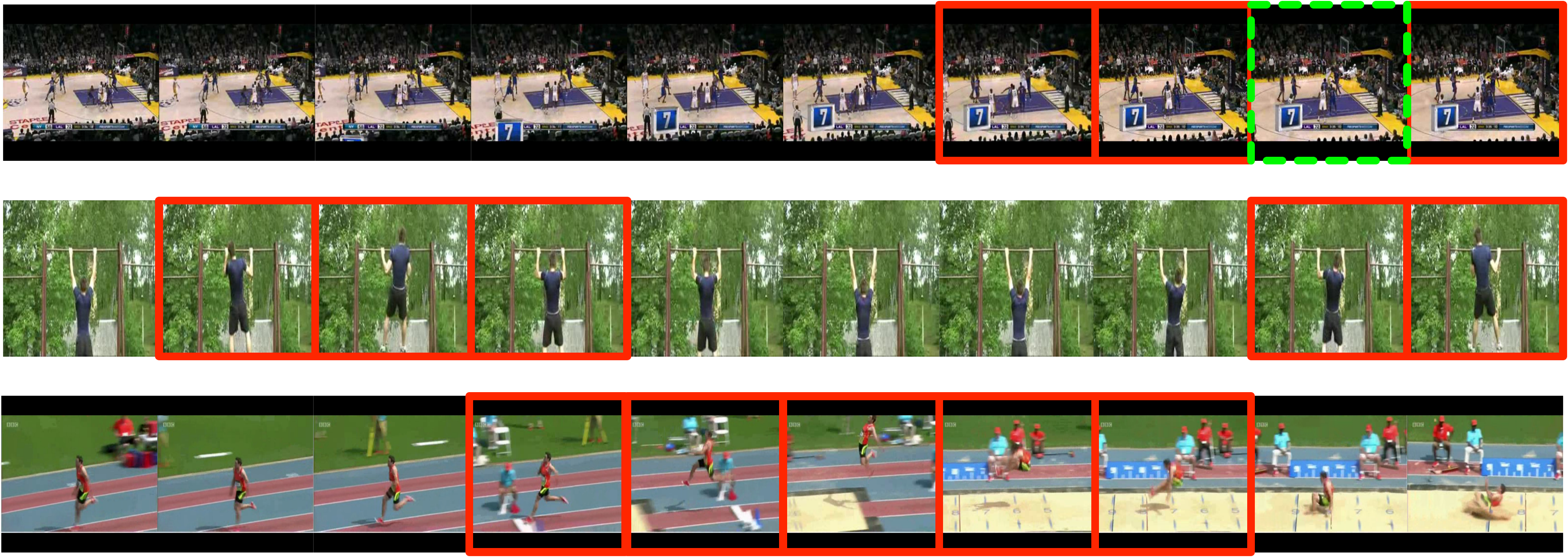}
\end{center}
\vspace{-6mm}
\caption{Illustration of the wrong predictions due to improper temporal attention. Top: Giving a low temporal weight for the important frame may mispredict the action ``Dunk" as ``Shot basketball". Middle and bottom: Applying the temporal attention learned from a seen action ``pull up" to an unseen action ``long jump" may mispredict it as ``run". Frames with red boundary: important frames to make the prediction. Frames with green dot boundary: important frames but given low temporal weights.}
\label{fig:mismatch}
\end{figure*}

The video action recognition needs to consider spatial information in the frame and the temporal correlation among frames. Directly apply the deep 3D Convolution Neural Networks (CNNs)~\cite{carreira2017quo,tran2015learning,tran2018closer} or CNNs + Recurrent Neural Networks (RNNs)~\cite{ballas2015delving, donahue2015long} with few data lead to severe overfitting by training a complex model with insufficient data. Current few-shot action recognition approaches mainly focus on studying how to compare video descriptors. However, the specific design for preserving temporal information given few videos is neglected.

The first few-shot action recognition work~\cite{zhu2018compound} applied a simplified version of the self-attention manner to study the frame's temporal weights by calculating the dot product between frame representation and trainable variables. They generate the temporal weights only based on the individual frame contents but without considering adjacent frames. Thus the model may miss some critical frames during prediction. For example, the top video of Figure~\ref{fig:mismatch} has the label ``dunk basketball". However, the frame with the green dot boundary is given a low weight due to the overlapping of players. Therefore, the action may be predicted as ``shot basketball". Other works~\cite{zhang2020few, bo2020few} applied the temporal filters (1D convolution or Gaussian) to generate the temporal weights for different frames. However, the important scenes of various videos usually occur at different frames. Apply a temporal filter, which is learned from seen action, to novel action may lose important temporal information. For example, given two videos with ``pull up" and ``long jump" actions shown in the middle and bottom of Figure~\ref{fig:mismatch}. Apply the temporal filters learned from ``pull up" (the frame with red border) to ``long jump" clearly missing the ``jump" action. Therefore, the video may be predicted as ``run". Another branch of works~\cite{bishay2019tarn, cao2020few} directly calculated the frame-wise similarity. Clearly, these approaches cause a huge computational overhead. With these insights, we consider a good temporal attention approach for few-shot action recognition should able to: \textbf{1.} Generate ``continues" temporal weights for each frame. In a regular video, if a frame is important (e.g. contains the action), the frames near that frame are very likely also crucial since the action is continuous. Therefore, the temporal weights for these frames should be similar. \textbf{2.} Adjust the temporal weights for different videos. Different actions usually have different temporal patterns. Even for the same actions, the important scenes may still occur in different temporal positions.

In this paper, we propose a Contents and Length-based Temporal Attention (CLTA) to address the few-shot action recognition task. CLTA utilizes Gaussian likelihood functions to provide the temporal weights for individual frames. Different from previous works~\cite{zhang2020few, bo2020few}, CLTA trains two learning matrices to study a ``contribution" scores of mean and standard deviation from each frame representation, respectively. Then uses these scores to define the Gaussian. During the training, CLTA studies the ability to learn mean and standard deviation based on the video contents and length from the seen videos. Since the frame representations and length of various videos are usually different, CLTA can provide proper and customized temporal attention to the unseen videos. We evaluate our approach on UCF101, HMDB51 and Kinetics. Our results show that CLTA outperforms other strong temporal attention methods (self-attention, TSF and SLDG) for few-shot action recognition by a large margin and achieves similar or better results compared to state-of-the-art approaches even without fine-tuning the backbone and a few-shot designed classifier. 

In summary, our main contributions are: \textbf{1.} a Contents and Length-based Temporal Attention (CLTA), a temporal attention framework for few-shot action recognition that can customize the temporal attention for unseen videos, which is often neglected in previous works. We use the soft-argmax to make our model fully differentiable. \textbf{2.} We conduct consistent comparative experiments to compare several representative temporal attention methods on the same testbed and show our approach outperforms other temporal attention methods by a large margin on all three datasets.

\section{Related Work}
\textbf{Action Recognition.} The CNN-based approaches have been widely applied to the action recognition area. Some works extend the CNNs to three-dimensional~\cite{carreira2017quo, ji20133d, tran2015learning, tran2018closer, varol2017long} to capture the spatio-temporal information of video. Other works still use 2D CNNs but process color and optical flow information in parallel for the subsequent late fusion of their separate classification scores~\cite{simonyan2014two, wang2015action, feichtenhofer2017spatiotemporal}. An alternative solution focuses on temporal rather than spatial information. The examples include model the temporal structure of video by various temporal pooling approaches~\cite{yue2015beyond, girdhar2017attentional}, rank functions~\cite{fernando2015modeling}, k-means clustering~\cite{girdhar2017actionvlad} and distribution functions~\cite{piergiovanni2018learning}. Recurrent Neural Networks have also been used to encode the temporal information for learning video representations~\cite{ballas2015delving, donahue2015long, srivastava2015unsupervised, sun2015temporal, sun2015human, wu2015modeling}. 

\textbf{Few-shot Learning of Image Classification.} Many efforts have been devoted to overcome the few-shot image classification problem. Some of the recent works address this by focusing on good model initialization~\cite{finn2017model, finn2018probabilistic, nichol2018first, rusu2018meta}. Therefore, when applying the classifier to predict novel classes, it can be learned with a limited number of labelled examples and a small number of gradient update steps. Another line of work focuses on learning an optimizer~\cite{ravi2016optimization, munkhdalai2017meta}. Examples include using the LSTM-based meta-learner to replace the stochastic gradient descent optimizer~\cite{ravi2016optimization} and applying a weight-update mechanism with external memory~\cite{munkhdalai2017meta}. These initialization-based methods can achieve rapid adaption with a limited number of training examples for novel classes. Another category focuses on similarity comparison. Researchers adopt component-wise distance~\cite{koch2015siamese}, cosine similarity~\cite{vinyals2016matching,gidaris2018dynamic,qi2018low}, Euclidean distance to class-mean representation~\cite{snell2017prototypical} and Graph Neural Network~\cite{garcia2017few} to measure the similarity between images.

\textbf{Few-shot Learning of Action Recognition.} There are only few works to address few-shot action reccognition problem. Zhu and Yang~\cite{zhu2018compound} propose the Compound Memory Network (CMN) which embeds frame features by multi-saliency function and predicts the class by comparing the dot product similarity of inputs. Bishay et al.~\cite{bishay2019tarn} calculate the relation between the query and support videos by measuring the similarity between aligned segments. Zhang et al~\cite{zhang2020few} train a 3D CNN with the self-supervised spatio-temporal mechanism to improve the robustness of the model and prevent over-fitting. Cao et al.~\cite{cao2020few} train the frame feature extractor by minimizing the frame-wise cosine distance between the support and query videos. 

Our work is similar to~\cite{piergiovanni2018learning, bo2020few}. \cite{piergiovanni2018learning} utilizes Cauchy distributions as the template to generate temporal weights for frames. The centers and width parameters are defined by trainable parameters. \cite{bo2020few} applied Gaussian distributions to generate the temporal weights for action recognition. The mean and standard deviation are manually defined based on the length of videos. For each Gaussian, they introduced a trainable scale parameter. Both approaches share the temporal weights to various videos, thus may cause misprediction for unseen actions. In contrast, CLTA trains two learning matrices to study the center and width parameters instead of directly learning them. Therefore, our approach avoids sharing the same temporal weights to various videos. We discuss the detailed difference between CLTA with~\cite{piergiovanni2018learning, bo2020few} in Section~\ref{dis} and compare their performance in Section~\ref{exp}

\begin{figure*}[!th]
\begin{center}
\includegraphics[width=\textwidth]{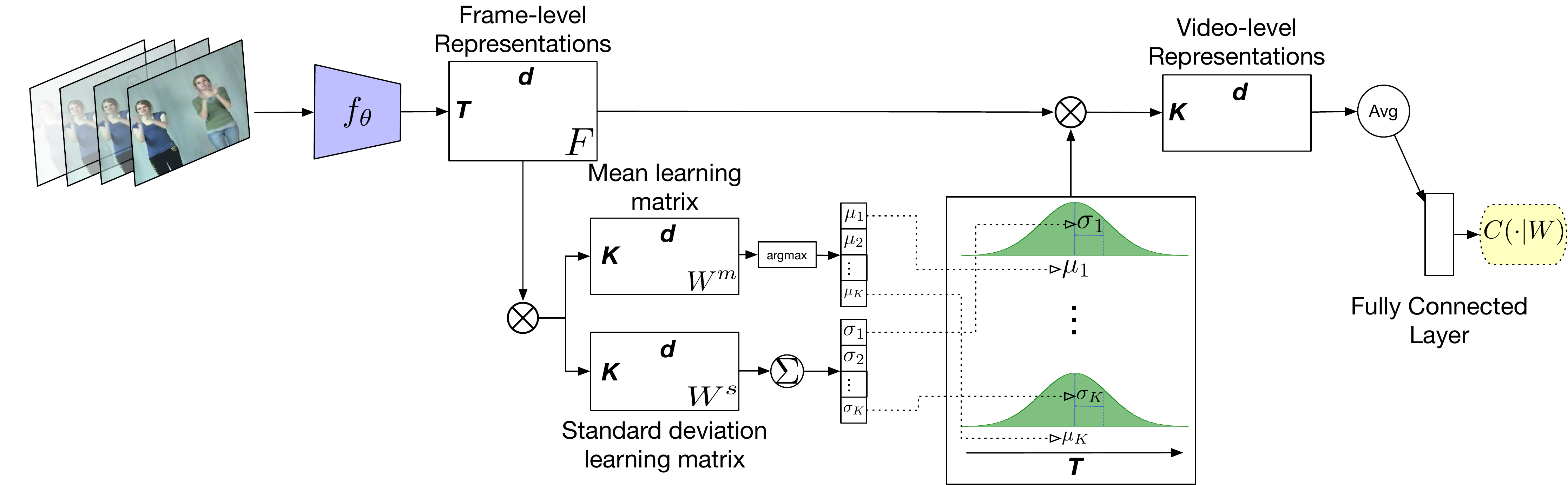}
\end{center}
\vspace{-6mm}
\caption{Overview of CLTA. $f_{\theta}$ is a 2D CNN which is used to extract frame representations. ``$\otimes$" denotes matrix multiplication and the proper matrix transpose is applied. ``$\Sigma$" denotes the addition operation according to the time dimension. $C(\cdot|W)$ is the classifier. Here, we show our approach with averaging video-level representations.}
\label{fig:model_architecture}
\end{figure*}

\section{Our Approach}\label{CLTA}
Given abundant labelled videos $X_b$ of base classes and few labelled videos $X_n$ of novel classes, our goal is training a model on $X_b$, which also could generate distinctive representations for novel classes in $X_n$. Therefore, a classifier is able to classify them with only few labelled videos. We achieve this by proposing a Contents and Length-based Temporal Attention (CLTA), the outline is shown in Figure~\ref{fig:model_architecture}. CLTA utilizes the Gaussian likelihood function to generate multiple temporal attentions for video frames since videos may have multiple crucial periods for classification. Instead of directly studying the mean and standard deviation, CLTA trains two learning matrices to study them based on the frame representations and video length. Therefore, CLTA can customize temporal attention and generate distinctive video representation for novel classes.

\textbf{Establish Frame-level Correlations via CLTA.}
Given the frame feature extractor $f_{\theta}$ (e.g. resNet), the frame-level representation matrix of a video is written as $F=(f_{\theta}(x_1), f_{\theta}(x_2), \ldots, f_{\theta}(x_T))$, where $f_{\theta}(x_t)\in \mathbb{R}^d$ is the representation of the $t^{th}$ frame. The $k^{th}$ temporal weights for the $t^{th}$ frame $a_{k,t}$ is defined by a Gaussian likelihood function as follows,
\begin{align}
    a_{k,t} &= exp(-\frac{1}{2}(\frac{t/Z-\mu_k}{\sigma_k})^2), \\
    \mu_k &= \frac{1}{Z}\argmax_t(f_{\theta}(x_1) \cdot w^m_k, \ldots, f_{\theta}(x_T) \cdot w^m_k), \label{eq_mean}\\
    \sigma_k &= \frac{1}{Z}\sum_{t=1}^T Sigmoid(f_{\theta}(x_t) \cdot w^s_k),
\end{align}
where $W^m, W^s \in \mathbb{R}^{K \times d}$ are the mean and standard deviation learning matrices and $W^m=(w^m_1, w^m_2, \ldots, w^m_K)$, $W^s=(w^s_1, w^s_2, \ldots, w^s_K)$. Here, the mean $\mu_k$ is the index of frame that $f_{\theta}(x_t) \cdot w^m_k$ achieves the maximum value. The standard deviation $\sigma_k$ is learned by calculating the sum of the dot products between $w^s_k$ and each frame-level representation $f_{\theta}(x_t)$. We apply the Sigmoid function to ensure the contribution from each frame is between 0 to 1. Both $\mu_k$ and $\sigma_k$ are then normalized by the maximum length of videos in the dataset $Z$ to preserve the video length difference. Since CLTA encode both video content (via dot product) and length (via summation) during study the mean and standard deviation of Gaussian, it is able to generate customized temporal attention for different videos (frame representations or length are different). The temporal attention $a_{k,t}$ is then normalized by a softmax function and used to aggregate the frame representations as the video-level representations. This could be formalized as
\begin{align}
    e_{k,t} &= \frac{exp(a_{k,t})}{\sum_{t=1}^Texp(a_{k,t})}, \\
    v_k &= \sum_{t=1}^T e_{k,t} f_{\theta}(x_t),
\end{align}
where $v_k$ stands for the $k^{th}$ video-level representation and $v_k \in \mathbb{R}^d$. 

\textbf{Video-level Representations Fusion.}
After we get $K$ video-level representations, each of them focuses on one important scene of the video. Before we make the prediction, these video-level representations need to be aggregated to form a single video descriptor $V$ that describes the entire video. The next question is how to fuse them? To be specific, we could treat every video-level representation equally important during the prediction by averaging them to get the video descriptor. $V = \frac{1}{K} \sum_{k=1}^K v_k.$ 

Another way is learning a soft-weight for each video-level representation, then calculating the weighted sum of them. $V =\sum_{k=1}^Ks_kv_k,$ where $s_k$ represents the soft-weight for the $k^{th}$ video-level representation.  The soft-weight is widely used in various video understanding tasks~\cite{piergiovanni2018learning, bo2020few}.

\textbf{Classifier.} 
The current Deep Neural Networks widely adopt Softmax classifier (a fully connected layer followed by a softmax function). The softmax classifier $C(\cdot|W)$ makes the prediction by calculating $W^TV \in \mathbb{R}^c$, where $c$ is the number of classes that needs to predict. 

The cosine distance-based classifier was introduced recently and has shown great performance for few-shot learning tasks~\cite{qi2018low, chen2019closer, cao2020few} since it can explicitly reduce the intra-class variations of data. The cosine distance classifier $C(\cdot|W')$ makes the prediction based on the cosine similarity scores $[s'_1, s'_2, \ldots, s'_c]$, where $c$ stands for the number of classes, $s'_i = V \cdot w'_i/\lVert V \rVert\lVert w'_i \rVert$ and $W' = [w'_1, w'_2, \ldots, w'_c]$. The prediction probability for each class is obtained by normalizing these similarity scores with a softmax function. Intuitively, the learned weights $[w'_1, \ldots ,w'_c]$ can be interpreted as prototypes (similar to~\cite{snell2017prototypical, vinyals2016matching}) for each class and the classification is based on the cosine similarity of the video descriptor to these learned prototypes. 

\textbf{Make CLTA Differentiable via Soft-argmax.}
While the CLTA approach is straightforward, the key technical challenge is that the argmax operation in equation~\ref{eq_mean} is not differentiable. Following the recent work for human pose estimation~\cite{luvizon2019human}, we use the soft-argmax with a scale parameter $\beta > 0$ to approximate the non-differentiable argmax operator in equation~\ref{eq_mean}.
\begin{align}
    argmax_t(f_{\theta}(x_1) &\cdot w^m_k, \ldots, f_{\theta}(x_t) \cdot w^m_k) \label{eq_softargmax} \\ &\approx \sum_{t=1}^T\frac{exp(\beta f_{\theta}(x_t) \cdot w^m_k)}{\sum_{i=1}^T exp(\beta f_{\theta}(x_i) \cdot w^m_k)}t. \nonumber 
\end{align} 
The use of soft-argmax in equation~\ref{eq_softargmax} helps the optimization process and allows gradients to be backpropagated through CLTA. 

\section{Discussion} \label{dis}
\textbf{Why use Gaussian?}
Self-attention, sometimes called intra-attention, is an attention mechanism relating different positions of a single sequence in order to compute a representation of the sequence. Self-attention has been used successfully in a variety of natural language tasks~\cite{parikh2016decomposable, paulus2017deep, vaswani2017attention}. A natural question is why using the Gaussian likelihood function generates the temporal weights instead of directly learning them by the learning matrices. We consider that in a regular video, if a frame is important (e.g. contains the action), the frames near that frame are very likely also important since the action is continuous. Therefore, the temporal weights for these frames should be similar. Self-attention does guarantee that the temporal weights for adjacent frames are similar. Consequently, some temporal information might be neglected. In contrast, using the Gaussian likelihood function could generate smooth temporal attention weights (similar weights are given to adjacent frames). We found that a smooth attention curve (generate by Gaussian) could better represent the action than a ``rough" curve (generate by learning matrices directly). We implement the self-attention as on one baseline and the detailed comparisons are shown in Section~\ref{exp} and Section~\ref{QRV}.

\textbf{Comparison to Similar Temporal Attention Approaches.} Some recent works also use probability likelihood functions to generate temporal attention for video understanding tasks~\cite{piergiovanni2018learning, bo2020few}. Temporal Structure Filter (TSF)~\cite{piergiovanni2018learning} trained multiple Cauchy likelihood functions as the templates to generate the temporal weights that are used to aggregate frame representations. Then they applied soft-attention to embed the aggregated frame representations. The centers and width of Cauchy are initialized by trainable and uniformly selected ``seeds" between $-1$ to $1$ then scaled according to the length of videos. TSF adjusts the Cauchy distributions by training the ``seeds". However, the Cauchy distributions are shared with various videos, which may cause it fails to capture some important scenes when applied to novel classes. 

Short-long Range Dynamic Gaussian (SLDG) applied the Gaussian likelihood function to generate the temporal weights for each frame. The mean and standard deviation are manually defined based on the length of videos. Then SLDG introduced an importance score learner to study the soft-weights to fuse the aggregated frame representations. SLDG still shares the same temporal weights to the video with the same length. The soft-weights only re-scale the Gaussian, but since they are still shared with different videos, the videos with the same length still have the same temporal weights. 

In few-shot learning, the aim is to classify novel classes with few training samples. Applying learned temporal weights for seen classes to novel classes may cause temporal information loss, thus making the wrong prediction. In contrast to~\cite{piergiovanni2018learning, bo2020few}, CLTA trains two learning matrices $W^m$ and $W^s$ to study the mean and standard deviation, respectively. During the study, both video contents and length are considered. In other words, CLTA study the ability of learning mean and standard deviation based on video contents and length during training. Therefore, it can be easily generalized to the novel classes. We implement both TSF and SLDG and the detailed comparisons are shown in Section~\ref{exp} and Section~\ref{QRV}

\section{Experiments}\label{exp}
\subsection{Datasets}\label{dataset}
We evaluate our approach on three popular datasets. First, UCF101~\cite{soomro2012ucf101} which consists of 13320 action videos in 101 categories. The second dataset is HMDB51~\cite{kuehne2011hmdb} which contains 6766 videos that have been annotated for 51 actions. For UCF101 and HMDB51 datasets, we follow the split as in~\cite{zhang2020few}. We select 70 classes as the training set, 10 classes as the validating set and last 21 classes as the testing set for UCF101 and 31 actions as the training set, 10 actions as the validating set and 10 actions as the testing set for HMDB51. We also use the Kinetics dataset~\cite{carreira2017quo} and follow the same split as in~\cite{zhu2018compound} which samples 64 classes for training, 12 classes for validation, and 24 classes for testing. Since some of the video are not available, we select other videos in the same class to guarantee each class will have $100$ videos. There are no overlap classes between the training, validating and testing sets for all these three datasets.

\subsection{Training and Testing Schemes} 
CLTA is first trained on the base class set $X_b$ (meta-training set). We choose the epoch which achieves the highest accuracy on the validating set. Then CLTA is evaluated on the novel class set $X_n$ (meta-testing set). During the testing phase, we construct the support set by randomly selecting $n$ classes from $X_n$, each of them contains $k$ randomly selected samples. The query set contains one sample from each of the $n$ classes. Therefore, each episode has a total of $n(k+1)$ examples. Samples in the query set and support set have no overlap with each other. We use the support set to only re-train a classifier since the high-way classifier is trained (e.g. $64$-way for Kinetics) during the training stage, which cannot be used directly for low-way classification (e.g. $5$-way). Then the classifier is used to classify the samples in the query set. The mean accuracies are reported by random sampling $10000$ episodes for all experiments.

\subsection{Implementation Details}
Follow the data augmentation as in~\cite{wang2016temporal}, we randomly crop from four corners and the center of input frames and sample the width and height of each crop randomly from $\{256, 224, 192, 168\}$, followed by re-sizing to $224 \times 224$. The argumentation is applied for both original and horizontal flipped frames. We use ImageNet pre-trained 152 layers ResNet~\cite{he2016deep} as the backbone for the experiments on UCF101 and HMDB51 and 50 layers ResNet for the experiments on Kinetics. For all experiments, we only use RGB frames as the input for the backbone.

The fully connected layer has dimension $1024$, followed by a ReLU function and a batch normalization layer. We also apply a dropout layer with the dropout rate $0.9$ to introduce a strong regularization. The scale parameter $\beta$ in equation~\ref{eq_softargmax} is set to $1e3$. During the training phase, we fix the backbone parameters and only train CLTA by minimizing the standard cross-entropy classification loss using Adam optimizer. The initial learning rate is set to $0.0001$ for UCF$101$ and HMDB$51$ and $0.001$ for Kinetics. The learning rate decays every $5$ epochs by $0.5$. We train CLTA with batch size $128$. In the testing phase, we fix the parameters of both backbone and CLTA then use the videos in support set to retrain a new classifier $100$ epochs with batch size $64$ and learning rate $0.001$ in each episode. The class scores of each augmentation are averaged as the final class score. 

\subsection{Ablation Study}\label{as}
\textbf{Number of Gaussian.}
We evaluate CLTA with different number of Gaussian. The softmax classifier is applied, and the results are shown in Table~\ref{tab:dc_num_G}. Increasing the number of Gaussian used in CLTA not necessary improve the performance. When the number is set to $6$, CLTA gives the best performances. We set the number of Gaussian to $6$ for all following experiments.
\begin{table}[!th]
\begin{center}
\resizebox{\linewidth}{!}{
\begin{tabular}{|cllllll|}
\hline
&\multicolumn{2}{c}{UCF$101$}& 
\multicolumn{2}{c}{HMDB$51$}& 
\multicolumn{2}{c|}{Kinetics-$100$} \\
\cline{2-7}
\# Gaussian & 1-shot & 5-shot & 1-shot & 5-shot & 1-shot & 5-shot\\ \hline \hline
3 & 77.0 & 86.4  & 57.0 & 75.4  & 66.8 & 83.4 \\
6 & \textbf{78.3} & \textbf{88.6}  & \textbf{58.7} & \textbf{76.8}   & \textbf{69.5} & \textbf{85.4} \\
9 & 76.8 & 86.7  & 56.5 & 75.2  & 66.5 & 83.2 \\
\hline
\end{tabular}}
\end{center}
\vspace{-6mm}
\caption{CLTA with different number of Gaussian (5-way accuracy).}
\label{tab:dc_num_G}
\end{table}
\begin{figure*}[!th]
     \begin{center}
     \begin{subfigure}[thbp]{0.49\textwidth}
         \centering
         \includegraphics[width=\linewidth]{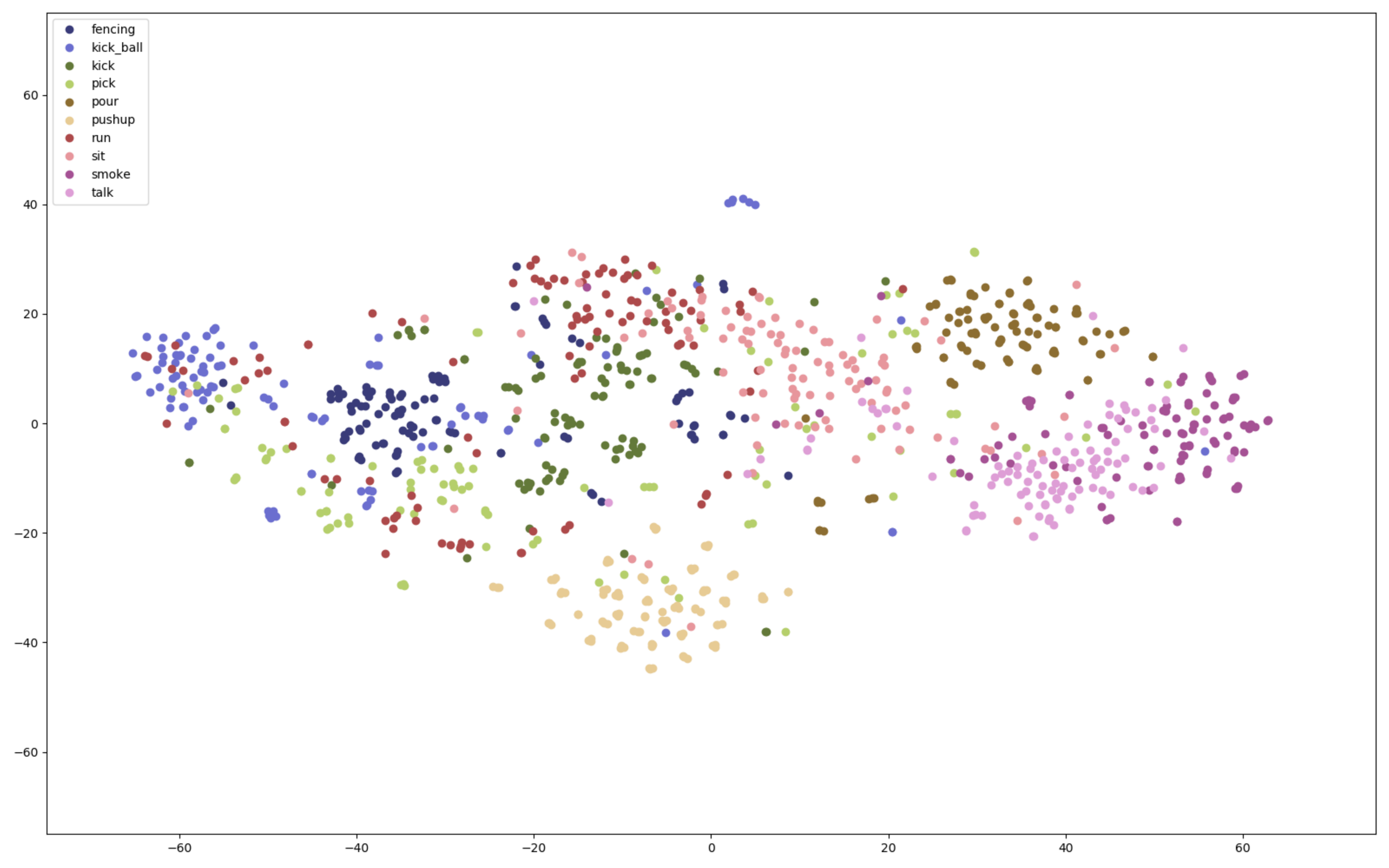}
         \caption{softmax classifier learned video representations}
         \label{fig:softmax}
     \end{subfigure}
     \begin{subfigure}[!th]{0.49\textwidth}
         \centering
         \includegraphics[width=\linewidth]{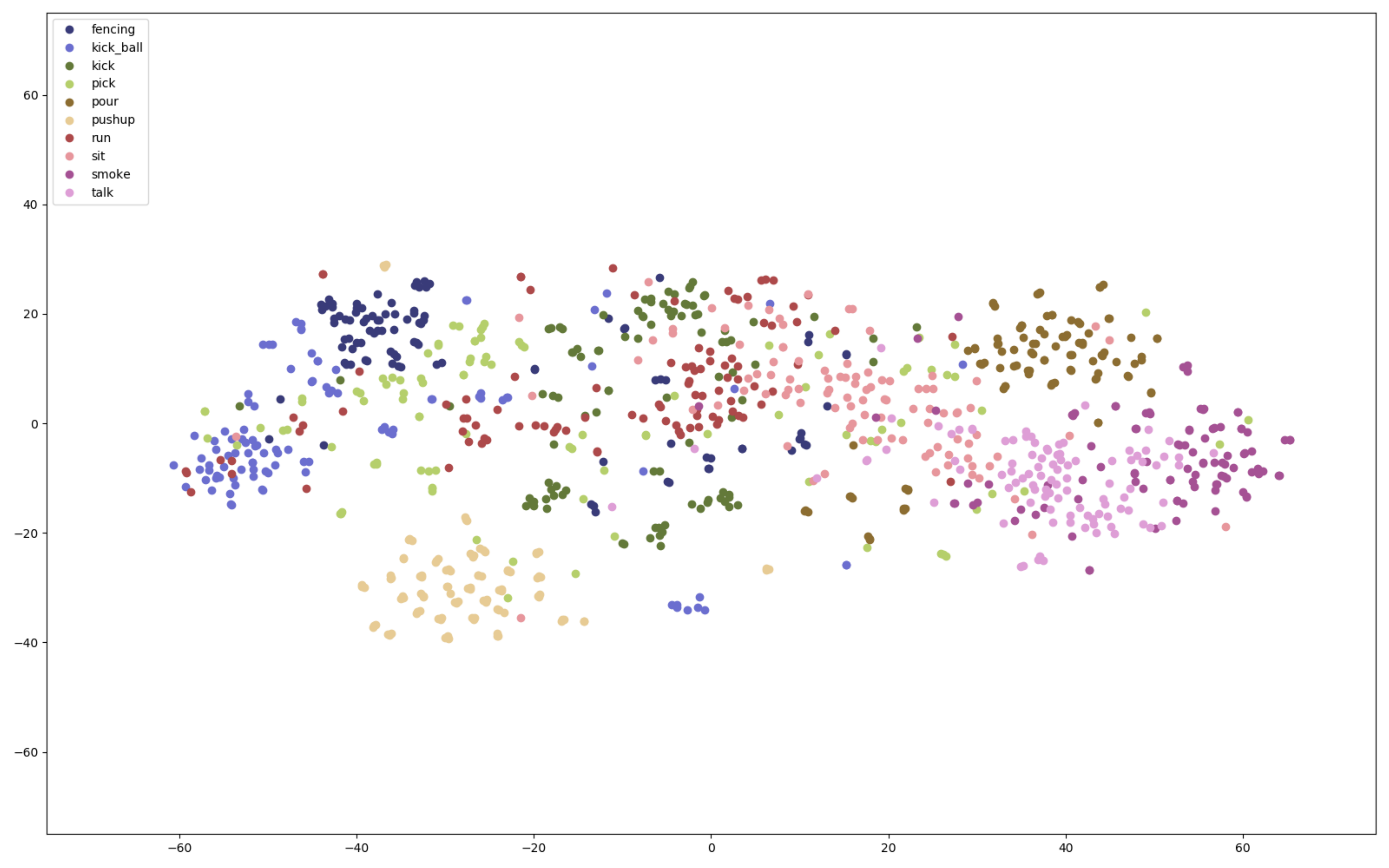}
         \caption{cosine distance classifier learned video representations}
         \label{fig:cos}
     \end{subfigure}
     \end{center}
     \vspace{-6mm}
\caption{We visualize the t-SNE scatter plots of the l2-normalized video representations learned with (a) CLTA with softmax classifier, and (b) CLTA with cosine distance classifier. The visualized video representations are from the test set of HMDB$51$.}
\label{fig:t-SNE}
\end{figure*}

\textbf{Video-level representations fusion.}
As mentioned in Section~\ref{CLTA}, CLTA generates $K$ video-level representations $\{v_1, v_2, \ldots, v_K\}$. Each of them highlights one important period of the video. Before making the prediction, these video-level representations need to be fused as a single descriptor $V$. We evaluate different fusion approaches and show the results in Table~\ref{tab:dc_vf}. Applying soft-weights to fuse the video-level representations not necessary improves the performance of CLTA. We believe that is because CLTA has already learned a good enough temporal attention which no need to be re-weighted.
\begin{table}[!th]
\begin{center}
\resizebox{\linewidth}{!}{
\begin{tabular}{|lllllll|}
\hline
&\multicolumn{2}{c}{UCF$101$}& 
\multicolumn{2}{c}{HMDB$51$}& 
\multicolumn{2}{c|}{Kinetics-$100$} \\
\cline{2-7}
Methods & 1-shot & 5-shot & 1-shot & 5-shot & 1-shot & 5-shot\\ \hline \hline
Soft-weight & \textbf{78.5} & 88.4  & 58.1 & 76.4  & \textbf{69.8} & 85.4 \\
Average & 78.3 & \textbf{88.6}  & \textbf{58.7} & \textbf{76.8}   & 69.5 & \textbf{85.4} \\
\hline
\end{tabular}}
\end{center}
\vspace{-6mm}
\caption{CLTA with different video-level representation fusion approaches (5-way accuracy).}
\label{tab:dc_vf}
\end{table}

\textbf{Classifiers.}
We evaluate CLTA with different classifiers and show the results in Table~\ref{tab:dc_classifier}. The cosine distance-based classifier have been shown a significant improvement compared to the ordinary softmax classifier for few-shot image classification~\cite{chen2019closer}. However, adopting cosine distance-based classifier for CLTA harms the performance (around 2\%). We also provide the t-SNE projection of video representations which are learned by two classifiers in Figure~\ref{fig:t-SNE}. The video representations studied by CLTA+cosine distance-based distance classifier are not shown more compaction compared to the video representations studied by CLTA+softmax classifier. The reason may be that the not fine-tuned backbone (pre-trained with softmax classifier) limits the ability of classifier to reduce intra-class variation.
\begin{table}[!th]
\begin{center}
\resizebox{\linewidth}{!}{
\begin{tabular}{|lllllll|}
\hline
&\multicolumn{2}{c}{UCF$101$}& 
\multicolumn{2}{c}{HMDB$51$}& 
\multicolumn{2}{c|}{Kinetics-$100$} \\
\cline{2-7}
Classifier & 1-shot & 5-shot & 1-shot & 5-shot & 1-shot & 5-shot\\ \hline \hline
Softmax & \textbf{78.3} & \textbf{88.6}  & \textbf{58.7} & \textbf{76.8}   & \textbf{69.5} & \textbf{85.4} \\ 
Cosine  & 76.9 & 87.1  & 56.8 & 75.8  & 68.2 & 84.1 \\
\hline
\end{tabular}}
\vspace{-6mm}
\end{center}
\caption{CLTA with different classifier (5-way accuracy).}
\label{tab:dc_classifier}
\end{table}

\textbf{Scale parameter $\beta$.}
We use soft-argmax to approximate the frame index $t$ that CLTA should give the highest temporal attention. Ideally, a large $\beta$ gives better approximation, but also might cause numerical errors in practice. We evaluate CLTA with different $\beta$ and show the results in Table~\ref{tab:dc_beta}. When setting $\beta=1e3$, CLTA achieves the highest accuracy. We also try to use a larger value (e.g. $\beta=1e4$), but it will cause Nan error during training.
\begin{table}[!th]
\begin{center}
\resizebox{\linewidth}{!}{
\begin{tabular}{|cllllll|}
\hline
&\multicolumn{2}{c}{UCF$101$}& 
\multicolumn{2}{c}{HMDB$51$}& 
\multicolumn{2}{c|}{Kinetics-$100$} \\
\cline{2-7}
$\beta$ & 1-shot & 5-shot & 1-shot & 5-shot & 1-shot & 5-shot\\ \hline \hline
1e1  & 76.2 & 86.1  & 55.7 & 74.9  & 66.5 & 82.1 \\
1e2  & 77.6 & 88.1  & 57.9 & 76.5  & 68.4 & 84.8 \\
1e3 & \textbf{78.3} & \textbf{88.6}  & \textbf{58.7} & \textbf{76.8}   & \textbf{69.5} & \textbf{85.4} \\ 
\hline
\end{tabular}}
\end{center}
\vspace{-6mm}
\caption{CLTA with different scale parameter $\beta$ (5-way accuracy).}
\label{tab:dc_beta}
\end{table}

\subsection{Demonstrate the effectiveness of CLTA}
\begin{table}[!th]
\begin{center}
\resizebox{\linewidth}{!}{
\begin{tabular}{|llll|}
\hline
\multicolumn{2}{|l}{}&
\multicolumn{2}{c|}{Kinetics-$100$} \\
\cline{3-4}
Methods & Temporal Attention & $1$-shot & $5$-shot    \\ \hline \hline
TSN~\cite{wang2016temporal}
 & Averaging            & 58.0 & 75.2 \\ 
TRN~\cite{zhou2018temporal} 
 & Multilayer Perceptron  & 61.7 & 76.1 \\
Self-attention
 & Trainable Matrix  & 65.7 & 79.5 \\ 
TSF~\cite{piergiovanni2018learning} 
 & Gaussian  & 63.9 & 77.2  \\ 
SLDG~\cite{bo2020few}  
 & Gaussian  & 64.2 & 78.1  \\ 
CLTA (ours) 
 & Trainable Matrices+Gaussian  & \textbf{69.5} & \textbf{85.4}\\
\hline
\end{tabular}}
\end{center}
\vspace{-6mm}
\caption{Demonstrate the effectiveness of CLTA for few-shot action recognition on Kinetics-100 (5-way accuracy). The Temporal Attention indicates how the temporal weights for each frame representation are learned. All approaches use ImageNet pre-trained resNet50 to extract frame-level representations. A softmax classifier is applied for this experiment.}
\label{tab:effectiveness_CLTA}
\end{table}

We compare our approach with several strong action recognition models to demonstrate the effectiveness of CLTA. The results are shown in Table~\ref{tab:effectiveness_CLTA}. We re-implement all these approaches and provide the same testbed for them. 

For TSN, we adopt the same sparse sampling strategy as introduced in~\cite{wang2016temporal}: splitting the video into $16$ equal-sized segments and randomly selecting one frame from each segment. This way, each video can be represented using a fixed length of frame feature sequences. The frame features are averaged as the video representation. For TRN~\cite{zhou2018temporal}, we use multilayer perceptrons to fuse frame features of different frames as the video representation instead of averaging. For self-attention, TSF and SLDG, we use all frame features as the input since they all can handle videos with various lengths. The only difference compared to CLTA is that we use a single learning matrix $W =(w_1, w_2, \ldots, w_K) \in \mathbb{R}^{K \times d}$ to generate the temporal weights for each frame by $f_{\theta}(x_t) \cdot w^k$. $K$ is also set to $6$ in this experiment. For TSF, we adopt the Gaussian likelihood function instead of Cauchy to guarantee that it is evaluated in the same testbed and follow all other settings mentioned in~\cite{piergiovanni2018learning}. For SLDG, we use the default setting as described in~\cite{bo2020few}. 

We have the following observation by comparing the evaluation results in Table~\ref{tab:effectiveness_CLTA}. \textbf{1.} If we take a close look at the temporal preserving part, the approaches without special few-shot temporal preserving design outperform the approaches that do not consider temporal information preservation (TRN, TSF and SLDG vs TSN). However, they are inferior compared to the approaches with special design for few-shot learning (TSF and SLDG vs self-attention and CLTA). The reason is that TSF and SLDG shared the learned temporal filters from the seen actions to unseen actions. Therefore, the discriminative information of unseen classes might be filtered out. In contrast with them, self-attention and CLTA train the trainable matrices and use the matrices to generate the temporal attention which enables the model to produce different temporal weights to the unseen classes. In short, they have the ``learning to learn" ability. \textbf{2.} The models adopt Gaussian to generate temporal weights outperform not using Gaussian models (TSF, SLDG vs TRN, self-attention vs CLTA). This verifies a smooth temporal attention curve is better than a ``rough" curve as we discussed in Section~\ref{dis}. 

\subsection{Comparison to State-of-the-art}
\begin{table*}[!ht]
\begin{center}
\resizebox{.9\textwidth}{!}{
\begin{tabular}{|lllllllll|}
\hline
\multicolumn{3}{|l}{}&
\multicolumn{2}{c}{UCF$101$}& 
\multicolumn{2}{c}{HMDB$51$}& \multicolumn{2}{c|}{Kinetics-$100$} \\
\cline{4-9}
Methods &Fine-tune &Temporal Attention & $1$-shot      & $5$-shot     & $1$-shot      & $5$-shot & $1$-shot      & $5$-shot    \\ \hline \hline
Matching Net$^{\star}$~\cite{vinyals2016matching}   & N         & Averaging  &- &- &- &-        & 53.3     & 74.6     \\ 
MAML$^{\star}$~\cite{finn2017model}         & N         & Averaging  &- &- &- &-        & 54.2     & 75.3     \\ \hline
TSN++$^{\dagger}$~\cite{wang2016temporal}          & Y         & Averaging   &- &- &- &-       & 64.5     & 77.9     \\ 
TRN++$^{\dagger}$~\cite{zhou2018temporal}             & Y         & Multilayer Perceptron   &- &- &- &-             & 68.4     & 82.0     \\ \hline
CMN~\cite{zhu2018compound}        & N         & Multi-saliency  &- &- &- &-    & 60.5     & 78.9     \\
ARN~\cite{zhang2020few}           
 &Y &3D Conv &66.3 &83.1 &45.5 &60.6 &63.7 &82.4 \\
TARN~\cite{bishay2019tarn}           & N         & Temporal Alignment   &- &- &- &-          & 64.8     & 78.5     \\ 
CMN++$^{\dagger}$~\cite{zhu2018compound}              & Y         & Multi-saliency &- &- &- &-    & 65.4     & 78.8     \\ 
OTAM~\cite{cao2020few}               & Y         & Temporal Alignment   &- &- &- &-            & 73.0     & 85.8     \\ \hline

CLTA (ours) & N & Trainable Matrices+Gaussian & 78.3 & 88.6 & 58.7 & 76.8   &69.5 &85.4\\
\hline
\end{tabular}}
\end{center}
\vspace{-6mm}
\caption{Mean accuracy of 5-way action recognition on UCF101, HMDB51 and Kinetics-100. The Temporal Attention indicates how the temporal weights for each frame representation are learned. ARN uses 4-layer C3D conv blocks as the backbone and all other approaches use ImageNet pre-trained resNet50 as the backbone. CLTA uses resNet152 for UCF$101$ and HMDB$51$. The ``++" sign indicates that the model using cosine distance-based classifier and training procedure as described in~\cite{chen2019closer}. Fine-tune: end-to-end fine-tuning. $^{\star}$: Results from~\cite{zhu2018compound}. $^{\dagger}$: Results from~\cite{cao2020few}.}
\label{tab:ucf&hmdb&kinetics}
\end{table*}
We compared CLTA to classical few-shot image classification approaches including Matching Net~\cite{vinyals2016matching}, MAML~\cite{finn2017model}, strong action recognition model TSN~\cite{wang2016temporal}, TRN~\cite{zhou2018temporal} and state-of-the-art few-shot action recognition models including CMN~\cite{zhu2018compound}, ARN~\cite{zhang2020few} TARN~\cite{bishay2019tarn} and OTAM~\cite{cao2020few}. The results are shown in Table~\ref{tab:ucf&hmdb&kinetics}. 

By analyzing the results, we have the following observations: \textbf{1.} The methods that consider the temporal correlation (e.g. CMN, TARN, TRN++) between frames outperform those models that averaged the frame features (e.g. Matching Net, MAML, TSN++). This result is consistent with action recognition on large datasets. \textbf{2.} Training model with proper distance function (e.g. cosine distance) improves the model generalization on unseen classes (CMN vs CMN++), even the model is not designed for few-shot learning (TSN, TRN). \textbf{3.} CLTA outperforms all few-shot approaches which adopt a not end-to-end fine-tuned backbone by a large margin. The strong action recognition model (TRN++) is still inferior compared to CLTA even adopting a fine-tuned frame feature extractor. This indicates that a specially designed temporal information preserving approach is critical for few-shot action recognition, even more important than the good frame feature extractor. \textbf{4.} The state-of-the-art OTAM is superior to CLTA for $1$-shot action recognition. We believe there are two reasons: First, OTAM adopts a much stronger end-to-end fine-tuned backbone compared to CLTA. Second, the number of samples in the support set is insufficient to train a softmax classifier for $1$-shot learning since CLTA achieves the same level performance compared to OTAM for $5$-shot learning. On the other hand, TAM introduces a huge computational overhead since it compares the cosine distance frame-by-frame between the support and query videos to preserve temporal correlation among frames ($O(T^2)$ ops where $T$ is the video length). However, CLTA is much simpler ($O(T)$ ops).


\section{Qualitative Results and Visualizations}\label{QRV}
\begin{figure*}[!th]
\begin{center}
\includegraphics[width=\textwidth]{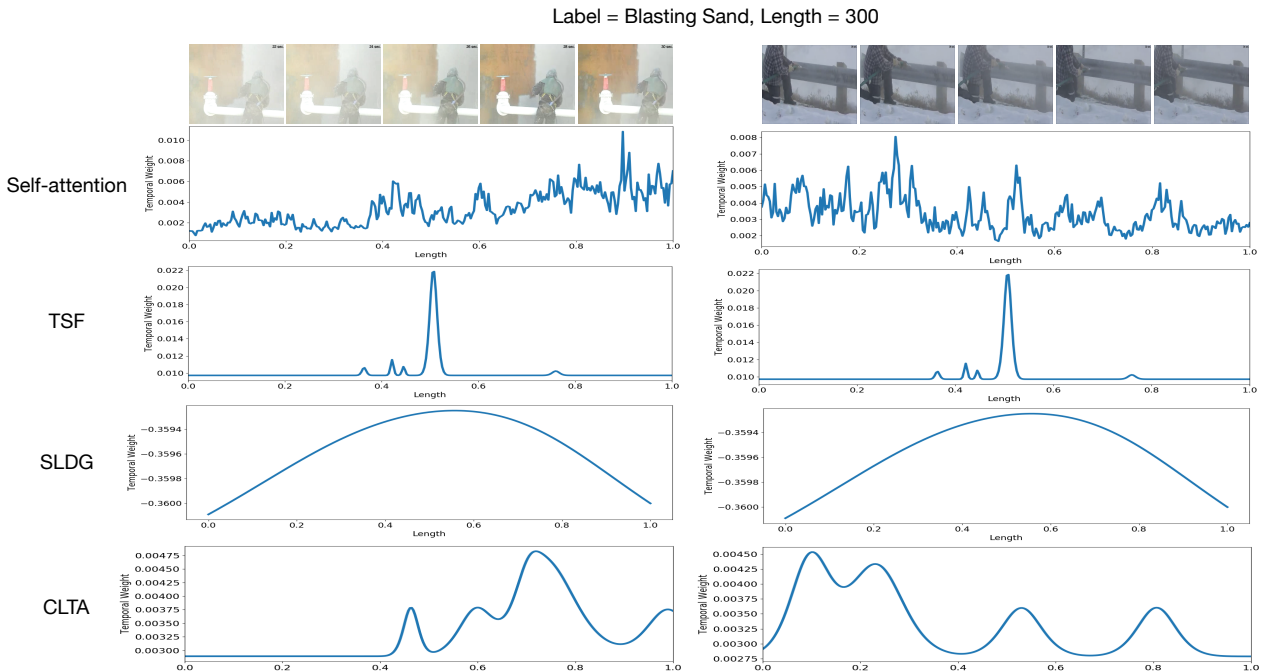}
\end{center}
\vspace{-6mm}
\caption{The visualization of temporal attentions that are learned by self-attention, TSF, SLDG and CLTA. The length of the video is mapped in the range [0,1]. We choose two sample videos with the same label and length from the Kinetics-100 test set to visually demonstrate the difference between these approaches. The softmax classifier applied during the training stage.}
\label{fig:visual_analysis}
\end{figure*}

We choose two videos with the same class and length as the examples and visualize of the learned temporal weights by self-attention, TSF, SLDG and CLTA in Figure~\ref{fig:visual_analysis}. 

All approaches can handle the videos with various lengths. As we discussed in Section~\ref{dis}, TSF and SLDG fail to adjust the temporal weights based on the content of unseen videos. Therefore, they apply the same filters, which are learned from the seen actions to the query actions. The self-attention is able to adjust the temporal attention for unseen videos. However, it gives sharp temporal weight curves, which may not correctly represent the temporal information of the video (e.g. For the left example, the 4th frame is more clear compared to other frames, but the self-attention model does not focus on it). In contrast with them, CLTA is able to provide smooth and customized temporal attention weights for unseen videos.

\section{Conclusion}
We propose a Contents and Length-based Temporal Attention (CLTA) for the few-shot action recognition task based on the idea various videos usually have different temporal patterns. Thus the learned temporal attention from seen actions cannot be directly applied to unseen actions. In contrast with previous approaches that study temporal attention directly, CLTA trains learning matrices to study temporal attention based on video contents and length. Therefore, CLTA can correctly capture the important periods of unseen videos. Our results show that CLTA outperforms other strong temporal attention methods (self-attention, TSF and SLDG) by a large margin for the few-shot action recognition task. CLTA also achieves similar or better results compared to state-of-the-art approaches even without using a fine-tuned backbone and a few-shot-designed classifier. 



\newpage

{\small
\bibliographystyle{ieee_fullname}
\bibliography{egpaper_final}
}

\end{document}